\documentclass[final]{cvpr}

\usepackage{times}
\usepackage{epsfig}
\usepackage{graphicx}
\usepackage{amsmath}
\usepackage{amssymb}

\usepackage[pagebackref=true,breaklinks=true,colorlinks,bookmarks=false]{hyperref}

\def\cvprPaperID{4} 
\def\confYear{CVPR 2021}

\begin{document}

\title{Cross-layer Attention Network for Fine-grained Visual Categorization}

\author{Ranran Huang, Yu Wang, Huazhong Yang \\
Tsinghua University\\
Beijing, China\\
{\tt\small hrr.rannie@gmail.com, \{yu-wang,yanghz\}@tsinghua.edu.cn}
\and
}

\maketitle

\begin{abstract}
Learning discriminative representations for subtle localized details
plays a significant role in Fine-grained Visual Categorization (FGVC). 
Compared to previous attention-based works, our work does not explicitly define or localize the part regions of interest; instead, we leverage the complementary properties of different stages of the network, and build a mutual refinement mechanism between the mid-level feature maps and the top-level feature map by our proposed Cross-layer Attention Network (CLAN). 
Specifically, CLAN is composed of 1) the Cross-layer Context Attention (CLCA) module, which enhances the global context information in the intermediate feature maps with the help of the top-level feature map, thereby improving the expressive power of the middle layers, and 2) the Cross-layer Spatial Attention (CLSA) module, which takes advantage of the local attention in the mid-level feature maps to boost the feature extraction of local regions at the top-level feature maps.
Experimental results show our approach achieves state-of-the-art on three publicly available fine-grained recognition datasets (CUB-200-2011, Stanford Cars and FGVC-Aircraft). Ablation studies and visualizations are provided to understand our approach.
Experimental results show our approach achieves state-of-the-art on three publicly available fine-grained recognition datasets (CUB-200-2011, Stanford Cars and FGVC-Aircraft). 
\end{abstract}

\section{Introduction}
Fine-grained Visual Categorization(FGVC) aims at identifying sub-categories of the same super-category, such as bird species \cite {wah2011caltech}, car models \cite{yang2015large} and aircraft types \cite{maji2013fine}. Compared to generic object recognition, distinguishing fine-grained objects remains a challenging task due to the subtle inter-category differences that mainly exist in local regions of objects. 
Therefore, the majority of works in fine-grained community focuses on learning more discriminative representations for subtle localized details. 


A significant line of previous works solve this problem by localizing discriminative attention parts with additional part annotations \cite{wei2016mask,zhang2014part} or only relying on image-level information~\cite{fu2017look,zheng2017learning,Wang_2018_CVPR,Liu2016Fully,Sun_2018_ECCV}.
Although impressive progress has been made, there are still several main drawbacks in such methods:
1) Collecting additional part annotations is costly. On the other hand, estimating accurate part locations without annotations often results in complicated pipelines.
2) The number of attention is pre-defined, which makes the model sub-optimal and less adaptive to new data.
3) With cropping the local region, the surrounding context is neglected, which restricts the expressive ability of extracted features.

Compared to those works, we consider capturing discrimination local attention without explicitly defining or localizing the attention regions; instead, we aim to exploit the relationships between mid-level feature maps and the top-level feature map, and automatically boost the feature extraction of local details at different stages of the network.

We first start from analyzing the properties of the feature maps at different stages of the network.
With sufficient reception field, the top-level feature map has richer global context information and encodes high-level semantic representations.
However, due to coarse spatial resolution, it tends to ignore fine-scale features at local regions, thus it is not effective to make fine-grained predictions only depending on top-level feature maps.
On the other hand, the mid-level feature maps contain richer local details and finer spatial information which are significant for identifying local regions in fine-grained datasets.
Nevertheless, the context information of the intermediate feature maps is limited, which may affect the semantic representation ability for object parts.
For instance, if the intermediate feature map responds to a small patch of feather in the original image, 
it is difficult for the network to identify whether this patch belongs to the wing or the chest of a bird. 
Therefore, the mid-level feature maps have finer spatial information but insufficient context information, while the top-level feature map contains richer context information but coarse spatial information.
Based on the the properties of the both, we propose to establish a mutual refinement mechanism between the middle feature maps and the top feature map. 
\begin{itemize}
    \item Refine mid-level feature maps with the context information of the top-level feature map.
    
    To enrich the context for each pixel of middle feature maps, we consider modeling rich contextual relationships over local features by leveraging the context information of the top feature map.
    To this end, we develop self-attention mechanism and design Cross-layer Context Attention (CLCA) module, which first computes the relations between different positions of mid-level feature maps using the combination of top-level and the mid-level feature map, then the relation matrix is used to perform self-attention on the mid-level feature maps.
In this way, the top-level feature maps help to provide richer and more accurate global context to make each position in mid-level feature maps have a better understanding of the surrounding pixels. 
    \item Refine the top-level feature map with the fine spatial information of mid-level feature maps.
    
As the mid-level layers contain fine spatial information on local part regions for recognizing fine-grained categories,  
we propose to build spatial attention maps from the middle feature maps to characterize the locations related to local parts, and 
highlight the response related to those local parts on the top-level feature map.
To be specific,
we propose Cross-layer Spatial Attention (CLSA) module, which utilizes the attention estimator generated by the mid-level feature map to impose spatial attention
on the feature map at the top level. 
In this way, the model is able to extract richer features of local regions from the top-level feature map.
\end{itemize}

\begin{figure*}[h]
\begin{center}
\includegraphics[width=0.9\linewidth]{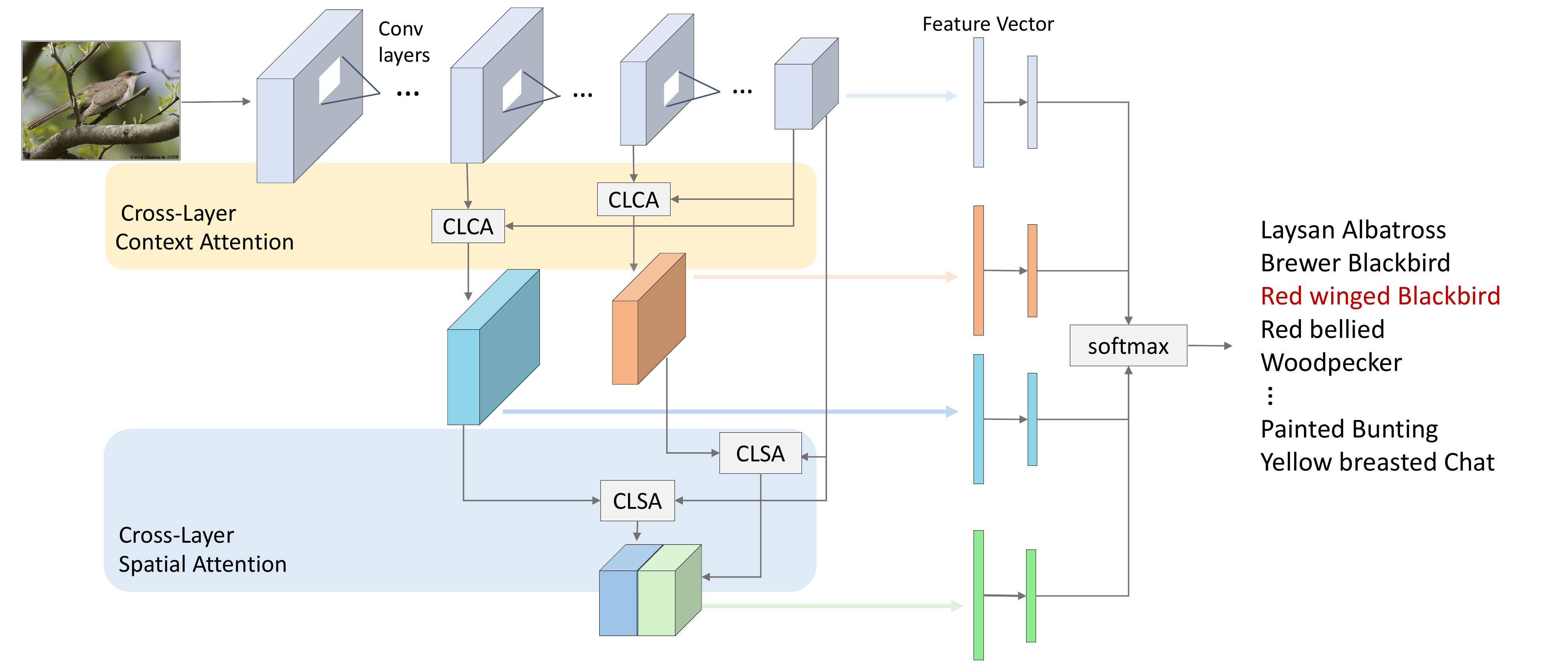}
\caption{Overview of our approach. The main components of our network are the Cross-layer Context Attention (CLCA) module and the Cross-layer Spatial Attention (CLSA) module. The CLCA module refines the mid-level feature maps by leveraging the top-level feature maps to enhance the global context information in the intermediate feature
maps.
The CLSA module takes advantage of the local attention in the mid-level feature maps to boost the feature extraction of local regions at the top-level feature maps.
Our final prediction can be obtained by combining the outputs at each branch. [Best viewed in color]}
\label{fig:framework}
\end{center}
\end{figure*}

Combining the CLCA module and CLSA module together forms our Cross-layer Attention Network (CLAN), which is presented in Figure \ref{fig:framework}. The main modules are depicted in Figure \ref{fig:clca_module} and Figure \ref{fig:clap_module}. Our contribution can be summarized into three aspects:
\begin{itemize}
    \item We propose a Cross-layer Context Attention (CLCA) module on the intermediate feature maps, which enhances the global context information and the representation ability for semantic parts of the intermediate feature maps.
    \item We introduce a Cross-layer Spatial Attention (CLSA) module on the top-level feature map to boost the spatial attention on local parts and improve the ability to extract local features at the top-level feature map. 
    \item Our experimental results demonstrate the effectiveness of our approach on three commonly-used fine-grained datasets without any additional annotations.

\end{itemize}

\section{Related Works}
\subsection{Fine-grained Visual Recognition}
Due to the small inter-class difference and high inner-class variations, how to identify the subtle differences in the crucial parts is of critical importance. A variety of methods have been developed to learn discriminative features.

B-CNN \cite{lin2015bilinear} is a symmetric two-branch CNN, of which the outputs are multiplied using the outer product to get second order statistics of convolutional feature maps. 
Image descriptors obtained in this way are capable of representing the pairwise correlations between feature channels, however, it is high-dimensional, which causes inconvenience in storage and subsequent processing. 
Therefore, improvements and extensions have been proposed by CBP \cite{gao2016compact} , LRBP \cite{kong2017low} and DBT-Net \cite{DBT-Net} with greatly reduced feature dimensions. 
The localization-based network is another stream to study fine-grained classification.
Earlier works of this category rely on additional part annotations to realize localization~\cite{wei2016mask,zhang2014part} by detection or segmentation schemes, and involve in heavy annotation burden. 
Thus, recently researches that only rely on image-level labels are drawing more attention~\cite{Liu2016Fully,fu2017look,zheng2017learning,Wang_2018_CVPR,Sun_2018_ECCV}. 
For instance, 
RACNN \cite{fu2017look}
propose a recurrent attention network to localize part regions from coarse to fine level, however, it only focuses on one single local part, and the alternative training process is non-trivial to optimize, moreover, parallel networks incur expensive computation cost.
To localize multiple object parts, MACNN \cite{zheng2017learning} proposes channel grouping loss to generate multiple parts by clustering. 
DFB-CNN \cite{Wang_2018_CVPR} learns discriminative patch detectors as a bank of convolutional filters, however, it needs a separate layer initialization to prevent the model learning from degeneration.
To exploit the inherent semantic correlations between part features, Sun \cite{Sun_2018_ECCV} proposes multi-attention multi-class constraint in the part learning process. 

Compared with previous works, our method also takes a weakly-supervised mechanism, yet, rather than learning bilinear feature representations or explicitly localizing attention regions, we leverage the properties of different feature maps to automatically exploit the local information using the proposed CLCA and CLSA module. 

\subsection{Visual Attention}
In the design of our CLCA module, we consider applying self-attention mechanism to improve the global context information in the mid-level feature maps. The self-attention module \cite{vaswani2017attention} is a method for machine translation that computes the response at a position in a sequence by taking the weighted average of all positions.
Xiaolong et al. \cite{wang2018non} bridges self-attention to non-local filtering means \cite{buades2005non} in computer vision, and gives a generic
non-local operation in deep neural networks. Non-local operations capture long-range dependencies directly by computing interactions between any two positions, and achieve good results on video classification.
We are inspired by their works.
Different from them, to help the earlier layers understand the global context better, we obtain the interaction matrix for positions of the input mid-level feature map using both the input feature map and the top-level
feature map.
This is because the global feature map contains more accurate global context information, and we hope this could provide the mid-level feature maps with a better
estimation of the relations between positions, and improve the context information of the middle feature maps.

For our proposed CLSA module, we are inspired by previous works \cite{senet, cbam} which generate attention estimator to refine the current feature map.
Different from them, we build attention in a cross-layer manner, \ie~leveraging the attention estimator generated by the previous layer to refine the top feature map, in which way we integrate the local details in the middle feature maps and the high-level semantic representations in the top feature map and generate a more robust feature representation for local regions.

\section{Proposed Method}
In this section, we will introduce our Cross-layer Attention Network (CLAN) presented in Figure \ref{fig:framework} by explaining the design of each component, \ie~Cross-layer Context Attention (CLCA) module and Cross-layer Spatial Attention (CLSA) module.
Suppose we leverage $N$ layers to make final predictions, of which the first $N-1$ layers belong to middle layers, and the last layer generates the top-level feature map.
 Let ${\bf L}^{s} = \{{\bf l}_{1}^{s},{\bf l}_{2}^{s}, \cdots, {\bf l}_{n_{s}}^{s}  \}$ be the feature map of a chosen middle layer $s \in \{1, \cdots, N-1\}$, where each ${\bf l}_{i}^{s}$ is the activation vector at the spatial position $i$ of $n_{s}$ total spatial
locations in the layer $s$, and ${\bf l}_{i}^{s} \in \mathbb{R}^{C_{s}}$ where $C_{s}$ denotes the the number of channels in layer $s$.
We denote the top-level feature map as ${\bf L}^{N} = \{{\bf l}_{1}^{N},{\bf l}_{2}^{N}, \cdots, {\bf l}_{n_{g}}^{N}  \}$, and $n_{g}$ is the spatial dimension, ${\bf l}_{i}^{N} \in \mathbb{R}^{C_{g}}$ where $C_{g}$ denotes the the number of channels of the top-level
feature map.

\subsection{Cross-layer Context Attention (CLCA)}
As we assume global spatial context is beneficial to enhance the representation ability of mid-level feature maps, the Cross-layer Context Attention (CLCA) module is designed and applied on the mid-level feature maps.
For each location $i$ in layer $s$, its relation with
location $j$ is denoted as $f({\bf l}_{i}^{s},{\bf l}_{j}^{s})$, $f$ is a pairwise function that outputs a
scalar relationship between the response at position $i$ and $j$ of the input signal.
The original self-attention computes the response at a position as a weighted sum of the features at all positions in the input feature maps,
thus enhance the capturing of long-range dependencies.

As top-level feature maps aggregate spatial information and have rich semantic information, we propose to leverage top-level
feature map to provide more accurate global context information. 
To match the global feature map with mid-level 
feature map, we upsample it to the spatial resolution $n_{s}$ of ${\bf L}^{s}$, In this way, we get $\hat{\bf L}^{N} = \{\hat{\bf l}_{1}^{N},\hat{\bf l}_{2}^{N}, \cdots, \hat{\bf l}_{n_{s}}^{N}  \}$
Then we fuse the mid-level feature map and the upsampled top-level feature map.
\begin{align}
      {\hat {\bf l}}_{i}^{s} = {\bf W}_{l}{\bf l}_{i}^{s}+{\bf W}_{g}{\hat{\bf l}}_{i}^{N}
\label{eq:fusion}
\end{align}
where ${\bf W}_{l} \in \mathbb{R}^{
C_{s}\times C_{s}}$ and ${\bf W}_{g} \in \mathbb{R}^{
C_{s}\times C_{g}}$.
We generate the interaction between position $i$ and $j$ using the combination of top-level features and mid-level features, so that the interaction could be more accurately estimated than solely using mid-level feature.
Then we use the interaction computed by the fused feature map to conduct a weighted sum of the response at all positions of the mid-level feature map. 
\begin{align}
    \hat{\bf y}_{i}^{s} = \frac{1}{C({\bf l })}\sum_{j}^{n_{s}}f(\hat{\bf l}_{i}^{s},\hat{\bf l}_{j}^{s})k({\bf l}_{j}^{s}) 
\label{cross-layer non-local}
\end{align}
Here $k({\bf l}_{j}^{s})={\bf W}_{k} {\bf l}_{j}^{s}$, ${\bf W}_{k} \in \mathbb{R}^{
C_{int}\times C_{s}}$.
Consider a residual unit, we get the final output $\hat{\bf z}_{i}^{s}$ as follows:
\begin{align}
    \hat{\bf l}_{i}^{s} = {\bf W}_{y}\hat{\bf y}_{i}^{s} + {\bf l}_{i}^{s}
\label{eq:clnl}
\end{align}
Here ${\bf W}_{y} \in \mathbb{R}^{
C_{s}\times C_{int}}$.
We call the process from Equation \ref{eq:fusion} to Equation \ref{eq:clnl} as a CLCA module, which can be simplified as Equation \ref{equation:clnl_processing} in terms of feature maps, and $\hat{\bf L}^{s}$ refers to the CLCA refined middle feature map. 
This CLCA module could be placed after the intermediate layer to help the mid-level feature map to capture more global context information and improve the representative ability for local regions.
\begin{align}
    \hat{\bf L}^{s} = CLCA({\bf L}^{s},{\bf L}^{N} ), s\in \{1,2,\dots,N-1\}
\label{equation:clnl_processing}
\end{align}

\subsubsection{Relation Metric}
For Equation \ref{cross-layer non-local}, we will discuss several widely used relation metrics $f$ in the following:

{\bf Gaussian:}
Gaussian uses dot product as the distance, then normalizes it with softmax operation.
\begin{align}
    f({\bf l}_{i},{\bf l}_{j}) &= e^{{\bf l}_{i}^{T}{\bf l}_{j}} , C({\bf l})=\sum_{j} f({\bf l}_{i},{\bf l}_{j})
\label{equation:gaussian}
\end{align}

{\bf Embedded Gaussian:}
Gaussian can be extended to Embedded Gaussian by computing similarity in the embedding space.
\begin{align}
    f({\bf l}_{i},{\bf l}_{j}) &= e^{\theta({\bf l}_{i})^{T}\phi({\bf l}_{j})} , C({\bf l})=\sum_{j} f({\bf l}_{i},{\bf l}_{j})
\label{equation:embedded_gaussian}
\end{align}
Where $\theta({\bf l}_{i}) = {\bf W}_{\theta} {\bf l}_{i}$, ${\bf W}_{\theta} \in \mathbb{R}^{
C_{int}\times C_{s}}$, $\phi({\bf l}_{j}) = {\bf W}_{\phi} {\bf l}_{j}$, ${\bf W}_{\phi} \in \mathbb{R}^{
C_{int}\times C_{s}}$. 

{\bf Dot Product:}
Another measure is to compute dot product in embedding space, and adopt the total location num of ${\bf l}$ as the normalization factor, which is denoted as $n$.
\begin{align}
    f({\bf l}_{i},{\bf l}_{j}) &= \theta({\bf l}_{i})^{T}\phi({\bf l}_{j}) , C({\bf l}) = n
\label{equation:dot_product}
\end{align}
Where $\theta({\bf l}_{i}) = {\bf W}_{\theta} {\bf l}_{i}$, ${\bf W}_{\theta} \in \mathbb{R}^{
C_{int}\times C_{s}}$, $\phi({\bf l}_{j}) = {\bf W}_{\phi} {\bf l}_{j}$, ${\bf W}_{\phi} \in \mathbb{R}^{
C_{int}\times C_{s}}$.

\begin{figure*}[h]
\begin{center}
\includegraphics[width=0.7\linewidth]{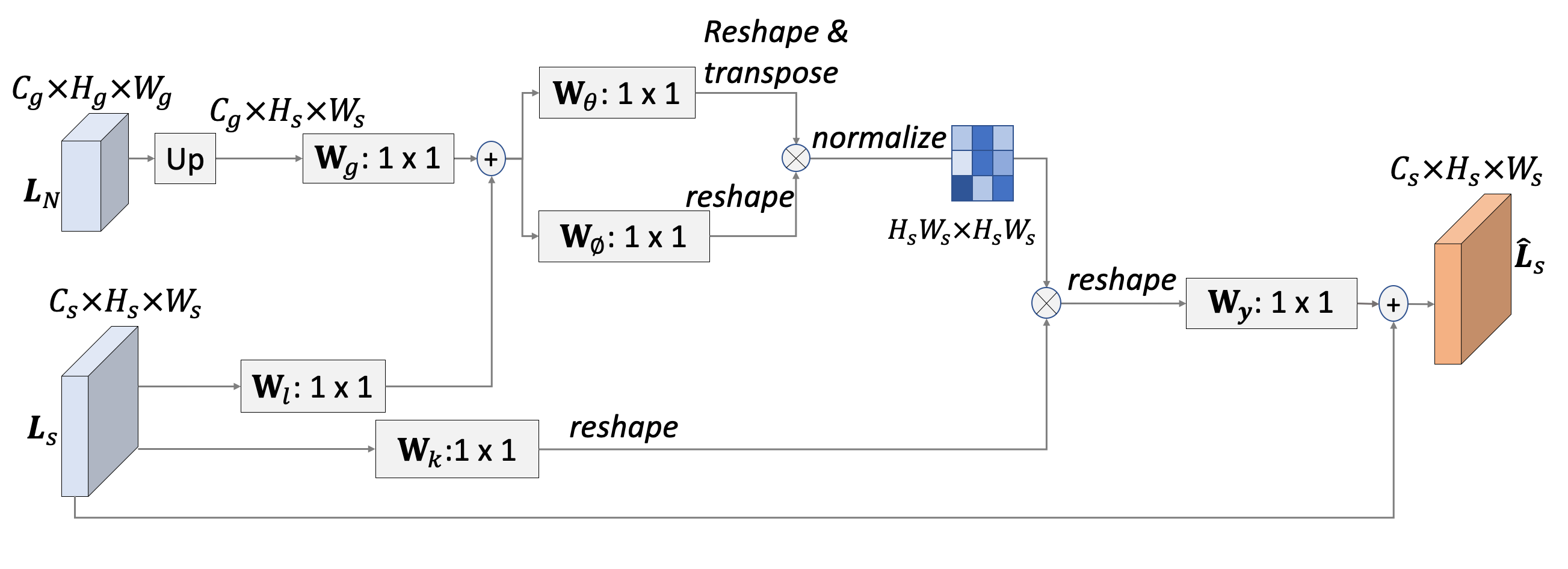}
\caption{The structure of Cross-layer Context Attention module, and ${\bigotimes}$
denotes matrix multiplication, ${\bigoplus}$ denotes element-wise sum.}
\label{fig:clca_module}
\end{center}
\end{figure*}

Take dot product as an example, the CLCA module can be implemented by the operation in Figure \ref{fig:clca_module}.
It can be observed that CLCA uses the combination of the top-level feature map and mid-level feature maps to estimate the interaction.

\subsection{Cross-layer Spatial Attention  (CLSA)}
Due to the refinement of CLCA module, the intermediate feature maps have more power to represent local regions, thus, we aim at leveraging the expressive ability of middle layers to enhance the local feature extraction of higher layers.
Specifically, we propose Cross-layer Spatial Attention (CLSA) module which is applied on the top-level layer and enforce spatial attention on the feature map of layer $N$ using the attention map generated by the middle feature map at layer $s$.

\begin{figure*}[h]
\begin{center}
\includegraphics[width=0.7\linewidth]{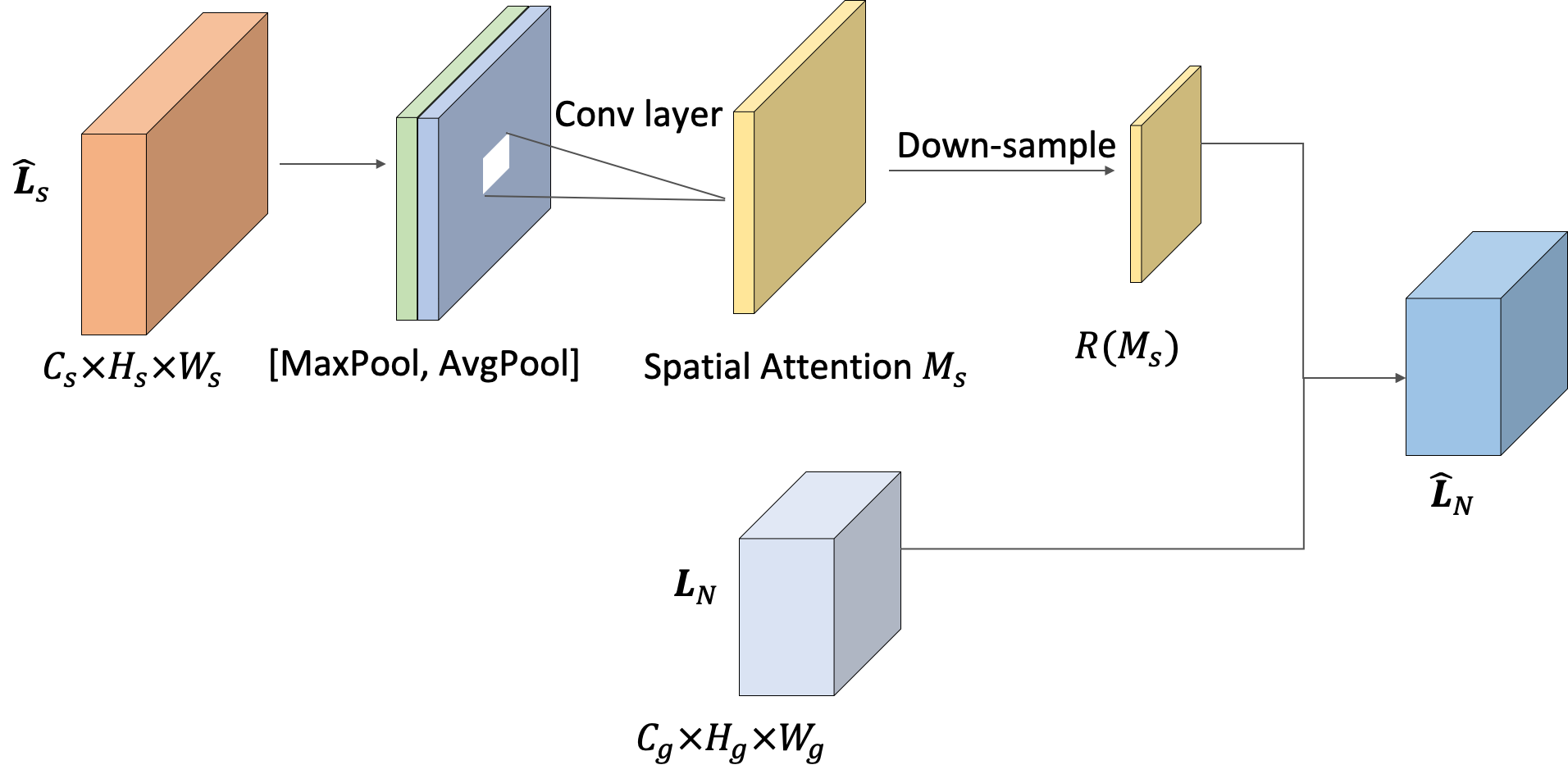}
\caption{The structure of CLSA module. Average pooling and max pooling operations are simutaneously applied on the CLCA refined feature map of layer $s$, then a convolutional layer is used to generate a spatial attention map, which is used to impose attention on the feature map of layer $N$. }
\label{fig:clap_module}
\end{center}
\end{figure*}

Inspired by \cite{cbam}, we first apply average pooling and max pooling operations simultaneously on the refined feature map of layer $s$ along the channel axis to highlight informative regions, then a convolutional layer is used to generate a spatial attention map ${\bf M}^{s}$. 
\begin{align}
\label{eq:mask}
    {\bf M}^{s} = K * ([AvgPool(\hat{\bf L}^{s});MaxPool(\hat{\bf L}^{s})]) 
\end{align}
Where $K$ are $3\times3$ kernels, and $*$ denotes convolution operation.
In order to integrate the property of finer local attention in the mid-level layers and the rich high-level semantic information in the top layer,
we enforce attention on the top layer with the attention estimator produced by the middle layer, which is shown in Equation \ref{eq:mask_weighted}.
\begin{align}
\label{eq:mask_weighted}
    {\bf L}^{s,N}  &= R({\bf M}^{s}){\bf L}^{N}, s\in \{1,2,\dots,N-1\}
\end{align}
Where $R$ denotes down-sample operation, which are used to transform the previous feature map to the spatial resolution of the top layer.
The process for Equation \ref{eq:mask} to Equation \ref{eq:mask_weighted} can be seen as a CLSA module, and simplified as Equation \ref{equation:clsa_processing}. The structure of CLSA can be depicted in Figure \ref{fig:clap_module}.
\begin{align}
    {\bf L}^{s,N} &= CLSA(\hat{\bf L}^{s},{\bf L}^{N} ), s\in \{1,2,\dots,N-1\} 
\label{equation:clsa_processing}
\end{align}

Finally we concatenate the top-level feature map attention pooled by different middle feature maps to get a refined top-level feature map.
\begin{align}
\hat{\bf L}^{N} &= [{\bf L}^{s,N};\dots;{\bf L}^{N-1,N}], s\in \{1,2,\dots,N-1\} 
\label{equation:feature_concat}
\end{align}

\subsection{Optimization}
In order to facilitate the learning of diverse and complementary intermediate feature maps, we use multi-scale feature maps over different spatial resolutions. 
The CLCA refined middle feature maps  (${\bf L}^{s}$) together with the original top-level feature map (${\bf L}^{N}$) and the CLSA refined top-level feature map ($\hat{\bf L}^{N})$) are sent to the pooling layer to get feature vectors, and then several separate fully connected layers are trained for feature at each scale to make independent predictions.

Our final prediction can be obtained by combining the outputs at each branch. And the objection function is the weighted sum of cross-entropy loss, as shown in Equation \ref{equation:loss}, and ${\bf y}$ denotes the labels of training samples.
\begin{align}
    loss &= \sum_{s=1}^{s=N-1}L(FC(Pool(\hat{\bf L}^{s})), {\bf y})  \\
    &+ L(FC(Pool({\bf L}^{N})), {\bf y}) \\
    &+ L(FC(Pool(\hat{\bf L}^{N})), {\bf y})
\label{equation:loss}
\end{align}

\section{Experiments}
\subsection{Experiment Setup}
{\bf Datasets}: Our experiments are conducted on three representative FGVC datasets:
 CUB-200-2011\cite{wah2011caltech} with 6k training images for 200 categories, Stanford Cars \cite{yang2015large} with 8k training images for 196 categories, and FGVC Aircraft \cite{maji2013fine} with 6k training images for 100 categories. 

{\bf Implementation Details:}
We use Pytorch framework to implement our experiments on GeForce GTX 1080 GPUs. 
The input image size of all experiments is $448\times448$, and no part or bounding box annotations are used in our experiments. 
The network structure of our approach is based on VGG-16~\cite{simonyan2014very}, ResNet-50~\cite{he2016deep} and ResNet-101~\cite{he2016deep}. 
We use the intermeditae feature maps generated by the fourth group (\ie~$conv4\_$x in \cite{he2016deep}) for ResNet-50 and ResNet-101. For VGG-16, we replace the last three fully connected layers by a global average pooling layer and a following fully connected layer, and consider the intermeditae feature map from layer $conv5\_2$ for VGG-16. 
We call this model setting as 2-scale CLAN, because the prediction depends on the top-level layer and one middle layer ($N=2$).
We also extend CLAN to three scales ($N=3$) based on 2-scale VGG-16 by involving in an extra intermediate layer $conv4\_3$. 
The network configuration details can be found in Appendix.

At initialization time, the base model layers are initialized from the ImageNet pretrained model.
In the experiment, the network is trained in an end-to-end manner, using an SGD optimizer with a momentum of 0.9, a batch size of 36, a weight decay of 1e-4, and a total training epoch of 120.
We set the initial learning rate to 0.01. In the experiment based on ResNet-50 and ResNet-101, the learning rate is decayed by 0.9 every two epochs; in the experiment based on VGG-16, the learning rate is decayed by 0.1 every 30 epochs.


\subsection{Ablation Studies}
{\bf Relation metric:}
Based on VGG-16, we first verify the effect of different relation metrics used in CLCA module on CUB-200-2011 \cite{wah2011caltech}. As shown in Table \ref{tabel:similarity_metric}, under 2-scale and 3-scale training, the effect of using the gaussian metric is worse than that of the embedded gaussian and dot product, while dot product and embedded gaussian can obtain similar results. For the sake of convenience, all the following experiments use dot product as the similarity measure of CLCA module.

\begin{table}[h]
\centering
\begin{tabular}{c|c|c}
\hline
Methods  & 2-scale & 3-scale  \\
\hline
Gaussian & 84.6 & 84.5   \\
\hline
Embedded Gaussian  & 85.1 & 85.7 \\
\hline
Dot Product  & 85.2 & 85.8   \\ 
\hline
\end{tabular}
\caption{The effect of different relation metrics of CLCA module on CUB-200-2011}
\label{tabel:similarity_metric}
\end{table}

{\bf Different methods applied on middle feature maps:} 
Based on 3-scale VGG-16, we compare the performance of different methods applied on the two middle feature maps.
According to Table \ref{tabel:clnl_nl}, GAP refers to the direct use of global average pooling in each intermediate feature map to obtain feature vectors and predictions; 
Non-local refers to the original non-local block which computes pixel relation only relying on the middle feature map, while CLCA module also introduces top-level feature map to generate the relation matrix.
Under the three methods, results from different layers are averaged to obtain the final prediction.

\begin{table}[h]
\footnotesize
\begin{center}
\begin{tabular}{c|c|c|c}
\hline
Methods & CUB & Cars & Aircraft   \\
\hline
GAP & 78.7 & 90.4 & 86.3 \\
\hline
Non-local & 84.8 & 92.5 & 89.5   \\
\hline
CLCA & {\bf 85.8} & {\bf 92.9} & {\bf 90.6} \\
\hline
\end{tabular}
\end{center}
\caption{The comparison of different methods applied on middle feature maps}
\label{tabel:clnl_nl}
\end{table}

It can be observed from the Table \ref{tabel:clnl_nl} that 
non-local and CLCA can greatly improve the accuracy, which shows that enriching context information do improve the representation power of the middle layer.
Compared with non-local, the CLCA module provides better result, which shows the top-level global information can provide a more accurate estimation of the correlation between positions of the middle-layer feature map, thus can help to boost the global context information in mid-level feature maps.
\begin{figure}[h]
\begin{center}
\includegraphics[width=1\linewidth]{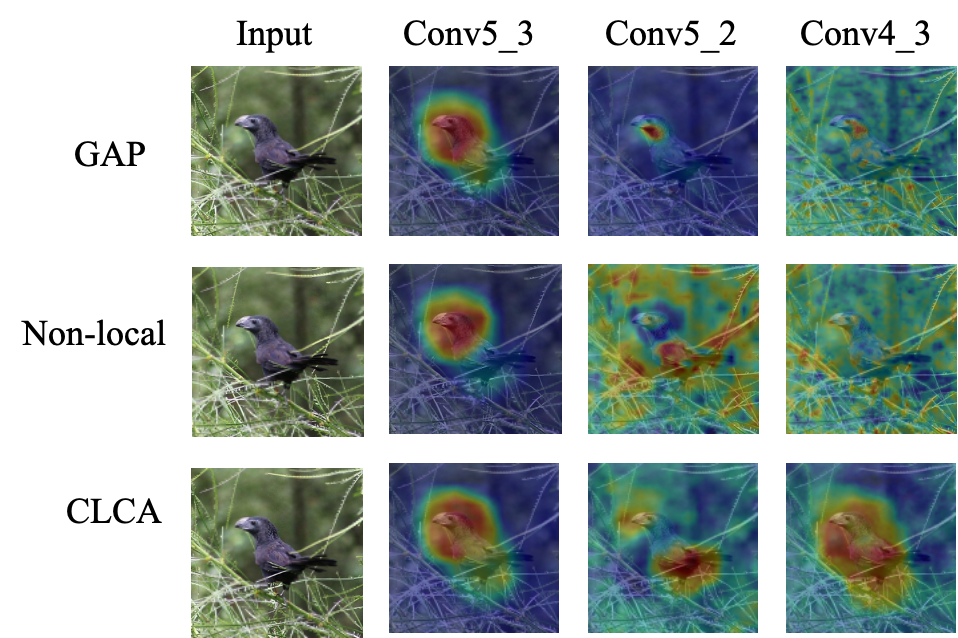} 
\caption{Visualizations of feature maps at each scale on different network configurations.[Best viewed in color]}
\label{fig:compare_clnl_nl}
\end{center}
\end{figure}

We also visualize the feature maps from each layer under three settings in Figure \ref{fig:compare_clnl_nl}. Compared to non-local, the middle feature maps of CLCA model is less reponsive to the background, and the attention is more centeralized on the objects.


\begin{figure*}[h!]
\begin{center}
\includegraphics[width=1\linewidth]{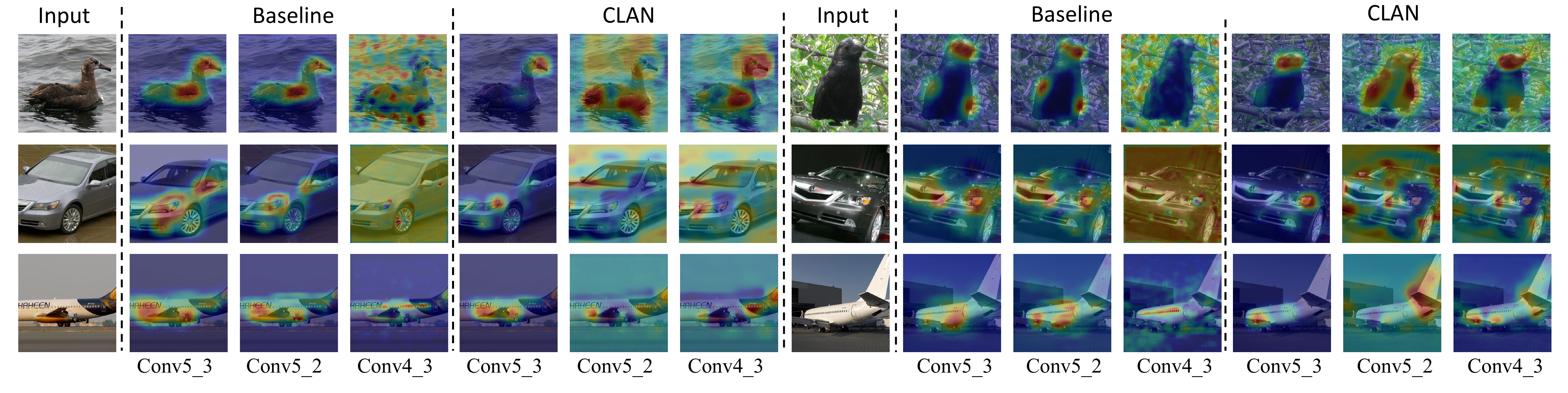}
\caption{Visualizations of feature maps at each scale on CUB-200-2011, Stanford Cars and FGVC Aircraft validation set. [Best viewed in color]}
\label{fig:compare}
\end{center}
\end{figure*}

\begin{figure}[h]
\begin{center}
\includegraphics[width=1\linewidth]{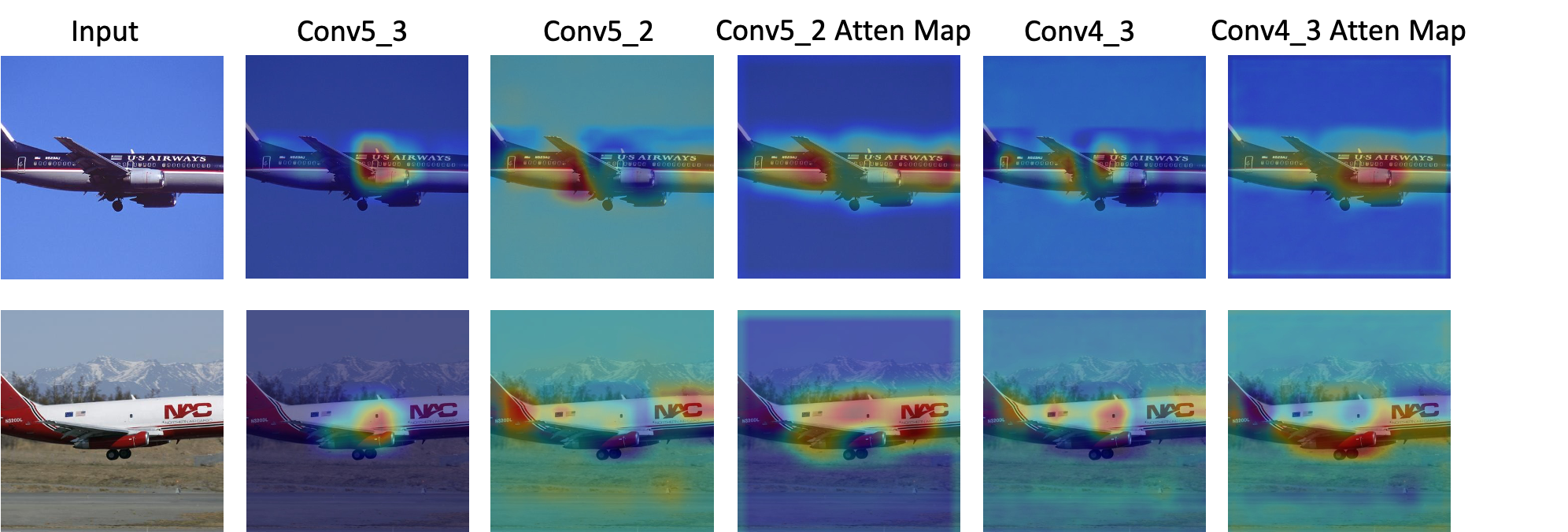}
\caption{The visualizations of CLSA attention maps on FGVC-Aircraft.}
\label{fig:clap_vis}
\end{center}
\end{figure}

{\bf Different configurations of CLSA module:}
Based on VGG-16, for CLSA, we first compare different pooling methods used during the process of generating attention map in Table \ref{tabel:clap_max_avg}, and verify applying average pooling together with max pooling works best.
\begin{table}[h]
\begin{center}
\begin{tabular}{cccc}
 \hline
          Methods & 2-scale & 3-scale  \\
          \hline
          CLCA + CLSA(Avg) & 85.3 & 85.6  \\
          \hline
          CLCA + CLSA(Max) & 85.7 & 86.0   \\
          \hline
          CLCA + CLSA(Avg\&Max) & {\bf 86.0} & {\bf 86.8}  \\
          \hline
 \end{tabular}
\end{center}
 \caption{The effect of different pooling methods used in CLSA on CUB-200-2011}
 \label{tabel:clap_max_avg}
\end{table}


{\bf The contribution of different streams:}
Based on 3-scale VGG-16, We show the impact of different streams on accuracy in Table \ref{tabel:branch_contribution}, where G-Branch, P-Branch and A-Branch refers to the classification branch corresponding to the $conv5\_3$, $conv5\_2$ and $conv4\_3$, respectively. CLSA-Branch is the prediction branch provided by CLSA module.
It can be observed that combining the global feature map and the middle feature maps can improve the performance, and multiple middle layers can further boost the accuracy, which shows that the middle layer makes up for the lack of local information in the global layer. And the CLSA branch can further obtain accuracy gains, indicating that refined by the middle feature maps, the top-level feature map can obtain richer feature representation for recognizing local parts of fine-grained data.
\begin{table}[h]
\small
\begin{center}
\begin{tabular}{c|c|c|c}
\hline
Branch  & CUB & Cars & Aircraft  \\
\hline
G-Branch  & 84.1 & 91.5 & 90.1   \\
\hline
P-Branch  & 85.2 & 91.8 &  90.3 \\
\hline
A-Branch  & 83.1 & 88.7  & 88.1 \\ 
\hline
CLSA-Branch  & 85.8 & 92.6  & 90.4 \\ 
\hline
G + P  & 85.3 & 92.1  & 90.6 \\ 
\hline
G + P + A & 85.8 & 92.9  & 90.6 \\ 
\hline
G + P + A + CLSA & {\bf 86.8} & {\bf 93.1}  & {\bf 91.0} \\ 
\hline
\end{tabular}
\label{tabel:branch_contribution}
\end{center}
\caption{The contribution of different streams at test time. Note that at training time a full CLAN model is trained, but the prediction only uses certain branchs}
\end{table}

\begin{table}[h!]
\begin{center}
\begin{tabular}{ccc}
\hline
Method & Params(M) & MACs(G)  \\
\hline
VGG-16  & 15.03 & 61.63\\
2-scale CLAN  & 16.08 & 62.30 \\
3-scale CLAN  & 17.34 & 64.83 \\ 
\hline
\end{tabular}
\end{center}
\caption{The comparison of complexity between baseline and CLAN at different scales based on VGG-16 }
\label{tabel:complexity}
\end{table}

\begin{table*}[h!]
\footnotesize
\begin{center}

\begin{tabular}{ccccccc}
\hline
Methods  & Base Model  & 1-stage & Sep. Init. & CUB & Cars & Aircraft\\
\hline
VGG-16~\cite{simonyan2014very} & VGG-16  & \checkmark & $\times$ & 79.1 & 87.0 & 85.1\\
ResNet-50~\cite{he2016deep} & ResNet-50  & \checkmark & $\times$ & 86.0 &92.0 & 89.9\\
ResNet-101~\cite{he2016deep} & ResNet-101  & \checkmark & $\times$ & 86.4 &92.5 & 90.5\\
\hline
B-CNN~\cite{lin2015bilinear}  & VGG-16  & \checkmark &  $\times$ & 84.1 & 91.3 & 84.1\\
CBP~\cite{gao2016compact}  & VGG-16  & \checkmark  &  $\times$ &84.0 & - &-\\
LRBP~\cite{kong2017low} & VGG-16  & \checkmark &  $\times$ &84.2 & 90.9& 87.3\\
DBT-Net~\cite{DBT-Net} & ResNet-50  & \checkmark  &   $\times$ & 87.5 & 94.1 & 91.2 \\ 
DBT-Net~\cite{DBT-Net} & ResNet-101  & \checkmark  &   $\times$ & 88.1 & 94.5 & 91.6 \\ 
\hline
RA-CNN~\cite{fu2017look} & VGG-19   & $\times$ &  \checkmark & 85.3 & 92.5& 88.2\\
MA-CNN~\cite{zheng2017learning} & VGG-19   & $\times$ &  $\times$ & 86.5 & 92.8&89.9\\
MAMC~\cite{Sun_2018_ECCV} & ResNet-50  & \checkmark &  $\times$ & 86.2 & 92.8&-\\
MAMC~\cite{Sun_2018_ECCV} & ResNet-101  & \checkmark &  $\times$ & 86.5 & 93.0&-\\
DFB-CNN~\cite{Wang_2018_CVPR} & ResNet-50 & \checkmark &  \checkmark & 87.4 & 93.8&92.0\\
TASN~\cite{TASN} & ResNet-50 & \checkmark  &   $\times$ & 87.9 & 93.8 & - \\
\hline
CLAN(2-scale) & VGG-16  & \checkmark  &   $\times$ & 85.9 & 92.9 &90.3\\
CLAN(3-scale) & VGG-16   & \checkmark  &   $\times$ & 86.8 & 93.1&91.0\\
CLAN(2-scale) & ResNet-50 & \checkmark  &   $\times$ & 88.1 &  94.2
 & 93.2 \\
CLAN(2-scale) & ResNet-101  & \checkmark  &   $\times$ & {\bf 88.7} & {\bf 94.5}
 & {\bf 93.3} \\
\hline
\end{tabular}
\label{table:fine_grained_result}
\end{center}
\caption{Results on CUB-200-2011, Stanford Cars and FGVC Aircraft. 1-Stage indicates the network is trained end-to-end after initialization. Sep. Init. denotes separate initialization}
\end{table*}

{\bf Complexity:}
We take CUB-200-2011 (the number of classes is 200) as an example, when the input image size is $448\times448$, we compare the difference in computational complexity between CLAN at two scales and the VGG-16 baseline. As can be seen from Table \ref{tabel:complexity} , 2-scale CLAN will only bring a slight increase in complexity. 
Compared to 2-scale CLAN, 3-scale CLAN processes an extra larger intermediate feature map,  thus brings further computation and parameter growth. Therefore, whether to leverage more intermediate layers is also a trade-off between accuracy and complexity.

As for previous fine-grained methods, it is notable that B-CNN~\cite{lin2015bilinear}, its variants (\eg~ CBP~\cite{gao2016compact}, LRBP~\cite{kong2017low}) and some multi-paralleled network (\eg~ RACNN\cite{fu2017look}) are usually not efficient to be deployed on resource-constrained platforms, because of their heavy computation and resource occupation. 
Thus our method is more efficient and only introduces two kinds of flexible modules, \ie~CLCA and CLSA module.

\subsection{Comparison with the state-of-the-art}
We compare our approach to a variety of previous weakly supervised fine-grained methods. These methods are sorted to three groups:

1) Finetuned baselines, including ResNet-50~\cite{he2016deep}, ResNet-101~\cite{he2016deep} and VGG-16~\cite{simonyan2014very}).  

2) End-to-end discrimination representation learning, including B-CNN~\cite{lin2015bilinear}, CBP~\cite{gao2016compact}, LRBP~\cite{kong2017low} and DBT-Net~\cite{DBT-Net}. 

3) Localization-based 
networks, including RACNN~ \cite{fu2017look}, MACNN \cite{zheng2017learning}, MAMC \cite{Sun_2018_ECCV}, DFB-CNN \cite{Wang_2018_CVPR}) and TSAN \cite{TASN}.

The results on CUB-200-2011, Stanford Cars and FGVC Aircraft are shown in Table \ref{table:fine_grained_result}. 
It can be observed that the proposed CLAN can outperform other previous works by a large margin.
Compared to MA-CNN which needs to be trained alternatively, RA-CNN which needs to pretrain the attention proposal
network and also takes an alternative training strategy, and DFB-CNN which requires a separate layer initialization in case the model converges to bad local minima, 
our approach only needs simple pretrained ImageNet initialization and could be trained end-to-end.

Based on VGG-16, 3-scale learning provides the accuracy of 86.8\%, 93.1\% and 91.0\% on three datasets respectively, which outperforms other VGG-based methods and also is a better result than 2-scale learning, and this
shows that feature maps at different scales could provide richer local information. 
Under the framework of ResNet-50, CLAN achieves the accuracy of 88.1\%, 94.2\% and 93.2\% on three datasets respectively, which outperforms previous ReNet-50 state-of-the-art result of TSAN~\cite{TASN}.
Under the framework of ResNet-101, CLAN achieves the accuracy of 88.7\%, 94.5\% and 93.2\% on three datasets respectively, which is better than DBT-Net~\cite{DBT-Net}.

\subsection{Visualization and Analysis}
Based on VGG-16, we visualize VGG-16 $conv5\_3$, $conv5\_2$ and $conv4\_3$ feature maps on CUB-200-2011, Stanford Cars and FGVC Aircraft validation set using algorithm in \cite{selvaraju2017grad}, and the results are shown in Figure \ref{fig:compare}.
We observe that $conv5\_3$ usually extracts the most salient
region of the bird, \ie~the head of birds.
For the baselines,
$conv5\_2$ pays attention to a similar area to the top layer $conv5\_3$, which makes the middle layer $conv5\_2$ unable to provide supplementary local information; the visualization of $conv4\_3$ indicates that the middle layer is more responsive to environmental noises.
For the visulization results of CLAN, 
activation maps from different layers concentrate on different regions of the objects, which indicates the intermediate layers could compensate the loss of local details in top-level feature maps.

Based on 3-scale VGG-16, we also visualize the attention maps at layer $conv5\_2$ and $conv4\_3$ generated by the CLSA module in Figure \ref{fig:clap_vis}. It can be observed that the attention map of the middle layer focuses on more diverse local information, 
therefore applying these attention maps to weight pool the top-level feature maps can integrate the high-level semantic representation and the diverse local attention, thus creating richer and more robust feature representations.

\section{Conclusions}
We proposed Cross-layer Attention Network (CLAN) to learn discriminative fine-grained features by building a mutual refinement mechanism between the mid-level feature maps and the top-level feature map. The core concepts of CLAN are CLCA and CLSA modules. Our proposed CLCA module leverages top-level feature maps to provide a more accurate global context for mid-level feature maps, and enhance the representation ability at the intermediate stage, and CLSA module boosts the local feature extraction at the top-level feature maps. Experiments evaluated on three benchmark datasets validate the effectiveness of our approach. Ablation studies further demonstrate the role of every component of CLAN.

{\small
\bibliographystyle{ieee_fullname}
\bibliography{egbib}
}

\end{document}